\documentclass[letterpaper, 10 pt, conference]{ieeeconf}  

\IEEEoverridecommandlockouts                              

\usepackage{graphics} 
\usepackage{epsfig} 
\usepackage{times} 
\usepackage{amsmath} 
\usepackage{amssymb}  
\usepackage[breaklinks, hidelinks]{hyperref}
\usepackage[capitalize]{cleveref}
\crefname{section}{Sec.}{Secs.}
\Crefname{section}{Section}{Sections}
\Crefname{table}{Table}{Tables}
\crefname{table}{Tab.}{Tabs.}

\title{\LARGE \bf
Improving Replay-Based Continual Semantic Segmentation with Smart Data Selection
}

\author{Tobias Kalb$^{1}$, Björn Mauthe$^{1}$ and J\"urgen Beyerer$^{2}$
\thanks{$^{1}$Tobias Kalb and Björn Mauthe are with Porsche Engineering Group GmbH, Porschestraße 911, 71287 Weissach, Germany
        {\tt\small \{tobias.kalb,
bjoern.mauthe\}@porsche-engineering.de}}%
\thanks{$^{2}$Jürgen Beyerer is with Fraunhofer Institute of Optronics, Systems Technologies and Image Exploitation IOSB and Vision and Fusion Laboratory (IES) of Karlsruhe Institute of Technology, 76131 Karlsruhe, Germany
        {\tt\small juergen.beyerer@iosb.fraunhofer.de}}%
}


\usepackage{url}
\usepackage{breakurl}

\begin{document}
\newcommand{\etal}{\textit{et al. }}

\maketitle
\thispagestyle{empty}
\pagestyle{empty}

\begin{abstract}

Continual learning for Semantic Segmentation (CSS) is a rapidly emerging field, in which the capabilities of the segmentation model are incrementally improved by learning new classes or new domains.
A central challenge in Continual Learning is overcoming the effects of catastrophic forgetting, which refers to the sudden drop in accuracy on previously learned tasks after the model is trained on new classes or domains.
In continual classification this challenge is often overcome by replaying a small selection of samples from previous tasks, however replay is rarely considered in CSS.
Therefore, we investigate the influences of various replay strategies for semantic segmentation and evaluate them in class- and domain-incremental settings.
Our findings suggest that in a class-incremental setting, it is critical to achieve a uniform distribution for the different classes in the buffer to avoid a bias towards newly learned classes.
In the domain-incremental setting, it is most effective to select buffer samples by uniformly sampling from the distribution of learned feature representations or by choosing samples with median entropy. Finally, we observe that the effective sampling methods help to decrease the representation shift significantly in early layers, which is a major cause of forgetting in domain-incremental learning.
\end{abstract}

\section{Introduction} \label{sec:intro}
A desideratum of many machine learning models exposed to a changing environment, is the ability to progressively acquire new knowledge without negatively interfering with previously learned knowledge. A major challenge in achieving this goal is to overcome catastrophic forgetting, where the model forgets knowledge learned from previous tasks while learning a new task \cite{McCloskey1989, French94catastrophicforgetting}.
This is especially relevant in constant changing environments, like automated driving, in which a model for semantic scene parsing has to be able to adapt to new unseen objects, e.g., e-scooters, or different driving situations. 
As it is intractable to collect representative training data that accounts for all possible driving scenarios from the beginning, a model starts with limited capabilities that are confined to very specific scope. This scope gradually increases over time, as new data are gathered.
Only recently, benchmarks for semantic segmentation were proposed to specifically address the challenge of class-incremental and domain-incremental semantic segmentation \cite{Tasar2019, Michieli2019, Klingner2020, Kalb2021}.
\begin{figure}[ht]
\centering
\includegraphics[width=\columnwidth]{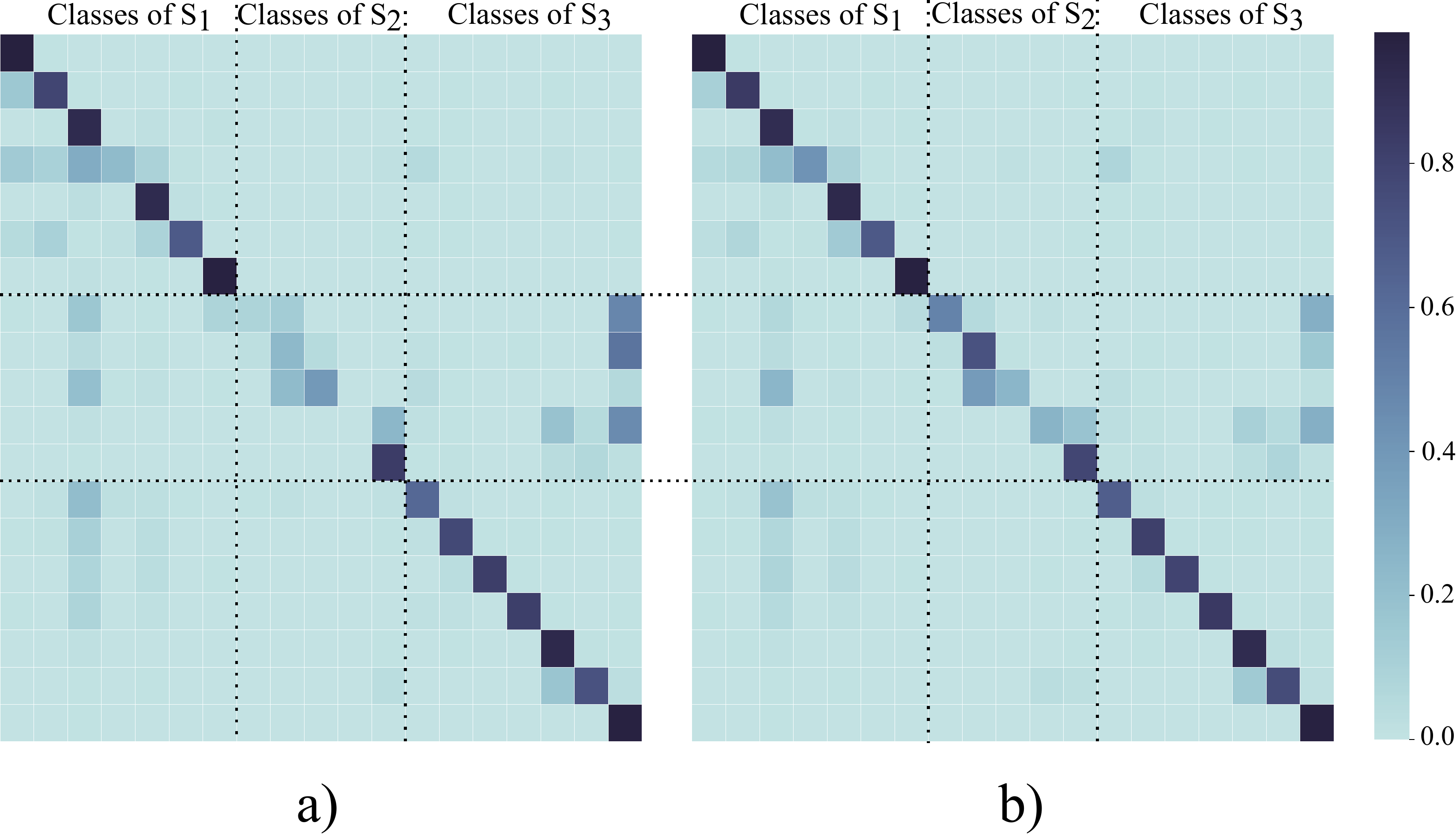}
\caption{Confusion Matrix after Class-Incremental Training on 3 Tasks of Disjoint-Cityscapes utilizing a) CIL \cite{Klingner2020} b) Class-Incremental Learning with Class-balanced Replay. Note the miss-classification for non-reoccurring classes of task $S_2$ for a) and the improved accuracy after employing class-balanced replay b).}
\label{fig:confmat_class_inc}
\end{figure}
\begin{figure}[ht]
\centering
\includegraphics[width=\columnwidth]{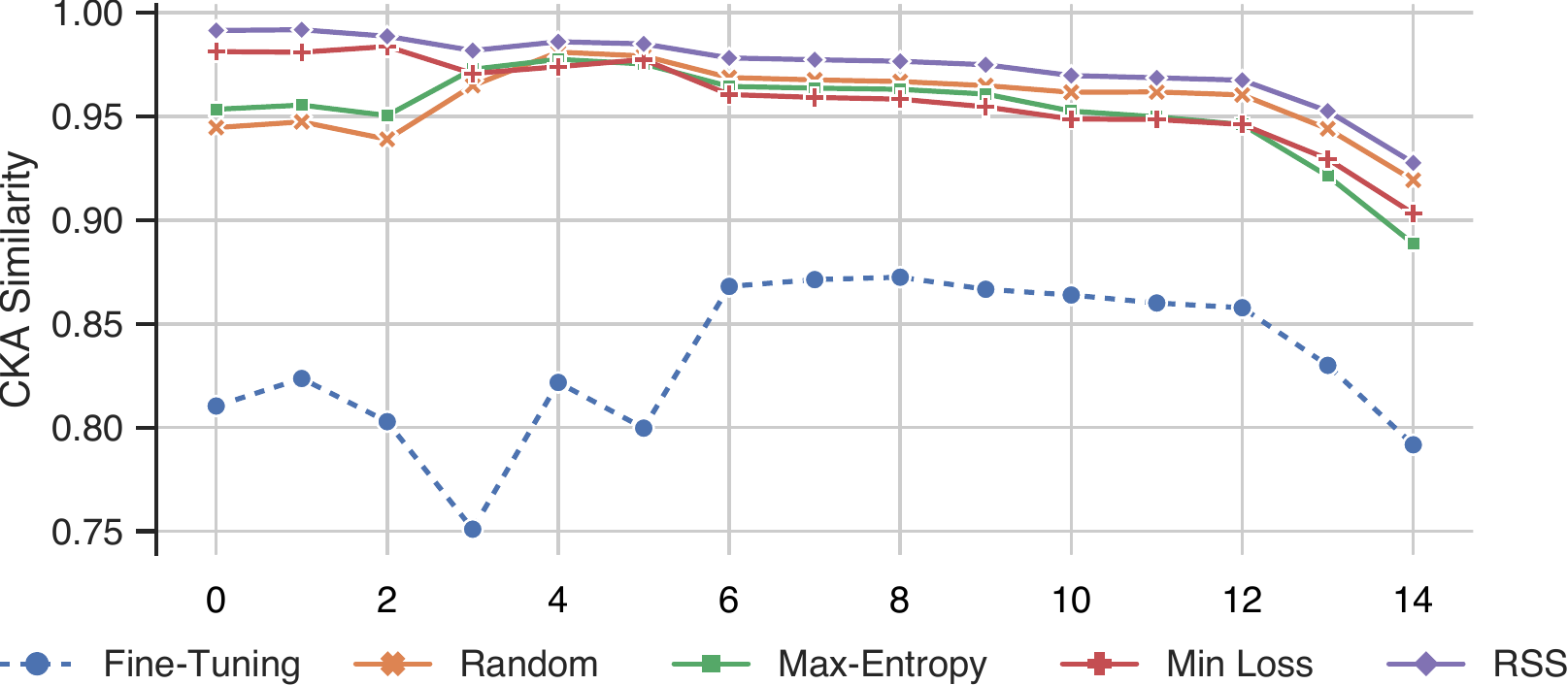}
\caption{Representational similarity (CKA) \cite{Kornblith2019} between activations of all layers (horizontal axis) before and after training on the second domain-incremental task. Solid Lines represent Replay-based methods with different sampling methods, the dotted line represents fine-tuning. Replay helps to stabilize the networks representations. An efficient sampling strategy can further improve the similarity which leads to less forgetting.}
\label{fig:cka_sim}
\end{figure}
However, a major difference between the proposed benchmarks of continual semantic segmentation and object classification, is that in semantic segmentation it is expected that old classes will reappear in the background of subsequent training images for new tasks. 
The nature of object classification prevents the reappearance of old classes in future training data.
We note that for semantic segmentation this assumption is justified in some applications, but it is unlikely to hold for every class.
Because of the selection of CSS benchmarks with recurring classes in the training images of subsequent tasks, most of the recently presented algorithms are purely based on knowledge distillation \cite{Tasar2019, Cermelli2020, Klingner2020, Douillard2020, Michieli2021}, which underperform in scenarios where classes will not reappear \cite{Kalb2021, Maracani_2021_ICCV}.
A simple solution to overcome this problem is to replay samples from previous tasks.
This has been shown to be effective in class- and domain-incremental scenarios \cite{Kalb2021, Maracani_2021_ICCV}.
However, in most cases the samples that are stored in the replay buffer are chosen randomly.
While random choice of samples has shown to be competitive with selective replay methods \cite{hayes2020remind, Aljundi.20.03.2019, Masana2020}, the performance of random choice can vary significantly, especially when learning in a sequence of class-unbalanced tasks. In this scenario, random choice can lead to a severe bias to only a few classes.
Therefore, we investigate various sample selection strategies for continual learning that will consistently outperform random sample selection for the task of semantic segmentation.\\
Our main contributions are:
\begin{enumerate}
    \setlength{\itemsep}{1pt}
    \setlength{\parskip}{0pt}
    \setlength{\parsep}{0pt}
    \setlength{\partopsep}{0pt}
    \item We adapt established sample selection strategies from classification to semantic segmentation and test them in class- and domain-incremental benchmarks and show that sample selection plays an important role when using replay in a continual learning setting, especially with strict memory limitations.
    \item We show that effective replay strategies help to stabilize the feature representations especially in deeper layers of the network during domain-incremental training. Therefore, methods that approximate the distribution of the internal representations of the task data, achieve the best results in our experiments.
    \item We demonstrate that in the class-incremental setting class-balancing is the most important property for sample selection, since it can effectively mitigate the effects of the task-recency-bias, observed in class-incremental learning.
\end{enumerate}

\section{Related Work}
\label{sec:relatedwork}
\subsection{Continual Learning}
Continual Learning is lately receiving more attention across many computer vision tasks such as classification \cite{Kirkpatrick2015, Aljundi2018, Aljundi.20.03.2019, hayes2020remind, NIPS2017_0efbe980, chenshen2018mergan}, object detection \cite{Acharya.14.08.2020, joseph2021open, Wang_2021_ICCV} and semantic segmentation \cite{Klingner2020, Michieli2019, Michieli2021, Kalb2021, Douillard2020}. 
Continual learning methods can be categorized in three main categories \cite{DeLange2019}: regularization-based methods, replay-based methods and dynamic architectures. 
Regularization-based approaches either constrain parameter updates on important parameters \cite{Kirkpatrick2015, Aljundi2018, pmlr-v70-zenke17a} or distill the knowledge from the previous model \cite{Li2018}. 
Replay-based approaches store a selection of the old training data for rehearsal, which is used during training on new data. 
Data can either be stored as raw images, internal representations of the model \cite{Acharya.14.08.2020, hayes2020remind} or in generative models like GANs or auto-encoders which approximate the previous data distribution \cite{NIPS2017_0efbe980, chenshen2018mergan, Zhai_2019_ICCV}. 
Finally, dynamic architectures adapt to new training data by dedicating a subset of parameters to each task by dynamically growing new branches \cite{AljundiExpertGate} or implicitly by masking parameters only for a specific tasks \cite{Cheung2019}.

\subsection{Incremental Learning for Semantic Segmentation}
In  class-incremental semantic segmentation, a new set of labels has to be learned in each new task while the previously learned classes remain unlabeled.
Therefore, the model has to learn to discriminate between old and new classes although they are never labeled within the same image.
This particular challenge is commonly addressed by adapting a knowledge distillation-based loss \cite{Hinton2015, Tasar2019}, in which a model trained on old data produces soft-labels for new training data in order to mitigate forgetting.
Recent work has improved this approach by stopping interference between old and newly introduced classes \cite{Michieli2019, Klingner2020} and by addressing semantic background shifts \cite{Cermelli2020,Douillard2020,Michieli2021}. 
However, these approaches require that previously learned classes reappear in the future training images, because otherwise they would still suffer from forgetting. Therefore, recent approaches integrate replay into incremental semantic segmentation either by replaying entire image-label pairs \cite{Maracani_2021_ICCV, Kalb2021}, by segments of each class \cite{Douillard2021} or by using a combination of pseudo-replay and out-of-distribution images for replay \cite{Maracani_2021_ICCV}.

\subsection{Replay-Selection Methods}
The importance of sample selection in continual learning has been discussed in various literature \cite{Castro.25.07.2018, Wu.30.05.2019, Acharya.14.08.2020, Aljundi.20.03.2019, Belouadah.2021}. 
Most Replay-based methods randomly choose samples to store in a buffer \cite{Chaudhry.02.12.2018, Ebrahimi.04.10.2020, LopezPaz.26.06.2017}, as it has proven to be a strong baseline \cite{Masana2020}.
Still, several methods deviate from this approach.
The method iCaRL by Rebuffi \etal \cite{Rebuffi.72017} uses a method called herding \cite{Welling.2009}, that selects the set of images for each class whose mean feature vector is closest to the actual mean feature vector of the class.
Aljundi \etal \cite{Aljundi.20.03.2019} proposed to select samples based on the diversity of the gradients to obtain a diverse set of samples. 
Chaudhry \etal \cite{Chaudhry.2018} investigated to select samples near the decision boundary assuming these to be more representative than those far away.
For object detection, Archarya \etal \cite{Acharya.14.08.2020} proposed selection methods that store samples with the highest or lowest number of different classes.

\section{Problem Formulation}
The task of semantic segmentation is to assign a class out of a set of pre-defined classes $\mathcal{C}$ to each pixel in a given image. 
A training task $T = \{(x_n, y_n)\}^{N}_{n=1}$ consists of a set of $N$ images $x\in \mathcal{X}$ with $\mathcal{X} = \mathbb{R}^{H \times W \times 3}$ and corresponding labels $y\in \mathcal{Y}$ with $\mathbb{Y} = \mathcal{C}^{H \times W}$. 
Given the task $T$ the goal is to learn a mapping $f_{\theta} : \mathcal{X} \mapsto \mathbb{R}^{H\times W\times |C|}$ from the image space $\mathcal{X}$ to a posterior probability vector $\hat{y}$. 
In the incremental learning setting the model $f_\theta$ is trained on a sequence of tasks $T_k$, that can introduce new classes or visually distinct instances from the same classes. 
The first setting is referred to as class-incremental setting, in this setting each task $T_k$ extends the previous set of classes $\mathcal{C}_{k-1}$ by a set of novel classes $\mathcal{S}_k$ resulting in the new label set $\mathcal{C}_k = \mathcal{C}_{k-1} \cup \mathcal{S}_k$. In this setting the labels of classes $C_{k-1}$ are not part of the training set of $T_k$.
The second learning scenario is domain-incremental learning, in which class set of subsequent tasks stay the same $\mathcal{C}_k = \mathcal{C}_{k-1}$, but the images are obtained from the different distributions, and therefore have distinct visual appearance.
When optimizing $f_\theta$ on the data of task $T_k$, the optimization disregards the previous task distributions $T_{i}$ with $i<k$, which leads to catastrophic forgetting of the previous task.
The main contributing factors of forgetting \cite{Masana2020} in CSS are:
\begin{itemize}
    \item Representation drift: While learning a new task, the network updates the internal feature representations to accommodate for new data. This is the main cause of forgetting in domain-incremental learning as the changing training distribution requires a change in representations, see \cref{fig:cka_sim}.
    \item Inter-task confusion: Since the classes are never jointly trained, the network does not directly learn to discriminate classes of different tasks. This applies especially to the class-incremental setting and can be alleviated using knowledge distillation. 
    \item Task-recency bias: In a class-incremental scenario the network has the tendency to miss-classify objects as belonging to classes of the latest task. 
\end{itemize}

\section{Sample Selection for Continual Semantic Segmentation}
\begin{figure}[h]
\centering
\includegraphics[width=\columnwidth]{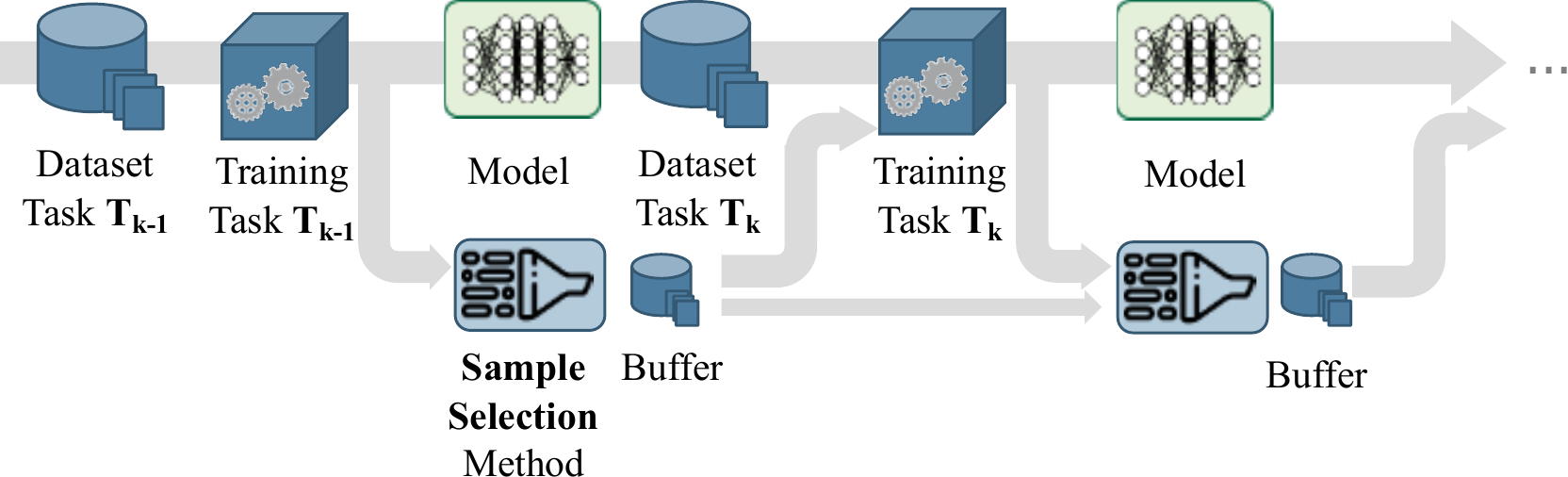} 
\caption{Visualization of the incremental learning process.}
\label{fig:replay_training}
\end{figure}

A common approach to overcome forgetting in continual learning is to replay samples from the previous tasks. 
However, a continual learning algorithm is constrained to a fixed memory budget that is not growing with the number of tasks. 
Therefore, rehearsal methods have to approximate the data distribution of the previous tasks $T_{k-1}$ by selecting representative samples that are stored in the memory $M$.
The procedure of replay-based continual learning and the sample selection is visualized in \cref{fig:replay_training}.
After training on a task $T_{k-1}$, we use a sample selection policy to sample data from the training dataset of $T_{k-1}$. 
This policy has access to all labels and images of $T_{k-1}$ as well as the currently trained model, but cannot access upcoming training data of $T_k$. 
The goal is to select samples to minimize forgetting of the task $T_{k-1}$ after training on subsequent tasks. 
The sampled data are stored in a buffer of a fixed size $M$ that is added to the training data of task $T_k$. 
If new data are added to the replay buffer when it is at maximum capacity, previously selected samples are discarded by the sample selection method.
Finally, a retrieval policy is used that samples data from the memory buffer during training. As we focus on investigating the selection policy, we use a fixed retrieval policy in all experiments that samples data uniformly over $M$.
Sample selection methods can broadly be divided in two categories: data-based selection methods that select data only based on the images and labels, and model-based sample selection that use outputs or internal activations of the model to select useful samples.

\subsection{Baselines Methods}
Random sampling provides a strong baseline for sample selection. Upon completing a task, random samples are chosen from the training data to be stored in the replay buffer. Once the buffer reaches its maximum capacity, random samples of the buffer are discarded and replaced with new samples.
In our experiments on random sample selection, we divide the memory buffer of size $M$ evenly between number of the already observed tasks $K$, so that there are $M / K$ samples in the buffer for each task.  
As multiple runs of random selection are expected to lead to deviating results, our goal is to develop a method that consistently outperforms the average random data selection.

\subsection{Data-based Sample Selection}
Data-based sample selection methods select samples purely based on the training data, without regard of the segmentation model. In our experiments we evaluate the following data-based sample selection:

\subsubsection*{Image-based sample Selection} 
The purpose of these methods is to ensure visually complex or diverse selection of samples, based on their visual appearance.
Our study compares three different methods: BRISQUE \cite{Mittal.2011}, Total Variation (TV) \cite{Rudin.1992}, and LPIPS \cite{Zhang.11.01.2018}.
BRISQUE calculates an image quality score with no reference, we calculate the quality score for every image in the dataset and then select the sample with the lowest score.
Total Variation is often used for denoising an image, as it measures the difference in luminosity between neighboring pixels. We use TV to capture more complex labels or images, containing either a lot of different classes or visually complex scenes.
The final approach in this category is to discard visual similar images from the buffer using LPIPS, we combine this approach with the selection of class balanced samples, named Div. Class-Balanced Samples.

\subsubsection*{Class-Balanced Sample Selection}
To ensure that individual classes are not forgotten, they need to be represented in the buffer. 
Therefore, the class-balanced sample selection methods tries to obtain a buffer with a uniform distribution of classes.
However, in semantic segmentation we expect that the classes within an image are rather unbalanced, e.g., the class \textit{street} in Cityscapes usually covers a large part of the image and classes like \textit{traffic light} only a very small part. 
At the same time some classes will appear in almost every image, e.g., \textit{Street} or \textit{Car}, while other classes like \textit{Train} are rare.
Therefore, we follow two different approaches to balance the classes in the buffer.
The first method, Class-Balanced Samples, selects samples that are closest to a uniform distribution of classes within an image. 
Class balanced buffer on the other hand greedily selects samples to steadily move the class-distribution in the buffer towards an uniform distribution of classes.
Ambivalent classes selects samples with the highest number of different classes to maximize the information content. 

\subsection{Model-based Sample Selection}
Model-based Sample Selection methods additionally use the outputs, embeddings or the gradients of the segmentation model as basis for the sample selection. The methods that are used in our evaluation are:

\subsubsection*{Loss-based Selection} 
This method selects samples based on the highest, lowest or median value of the cross-entropy loss.
The intuition is that the neural network is well adapted to frequently recurring situations in the dataset and has lower loss values on these samples, thereby we hypothesize that samples that are representative of the dataset distribution generally have a lower loss.
On the other hand, samples with the highest loss are expected to be not as representative for the data distribution, but might still be useful for replay due the high information content.
Finally, we also include a median loss selection, in order to avoid trivial training samples with low loss and outliers with a high loss.

\subsubsection*{Entropy-based selection}
In the entropy-based selection the prediction uncertainty is estimated using the Shannon entropy \cite{Shannon.1948}. 
Following the same intuition as in the loss-based selection, three methods are implemented selecting samples with the lowest, the highest, and samples close to the average uncertainty.

\subsubsection*{Gradient Based Sample Selection (GSS)} 
GSS selects the samples based on the diversity of the gradients \cite{Aljundi.20.03.2019}. 
In our experiments we use the greedy variant, which assigns a score $R_n$ to each sample based on the maximum cosine similarity of the gradients between the current sample and a fixed number of randomly selected buffer samples. 
The samples which have similar gradients get higher scores from GSS. Samples are added to the buffer with their respective scores. Once the buffer is full, the normalized score $R_n / \sum\nolimits_{m} R_m$ is used as the probability that a sample is discarded.
Contrary to Aljundi \etal, the available task boundaries are used and the process is performed for each task to ensure comparability across the methods. When adding a new task, the samples with the highest score are discarded to keep the samples with the highest divergence according to their gradients. 

\subsubsection*{Representation-based sample selection (RSS)} 
As recent work suggested that methods that mitigate forgetting have a stabilizing effect on the representations of deeper layers \cite{ramasesh2021anatomy}, we propose Representation-based sample selection (RSS). RSS selects buffer samples in a way to approximate the learned representations of the training data of previous tasks.
Therefore, we use the activations of each individual sample extracted by the encoder of the segmentation network and project them into a lower dimensional space, using UMAP \cite{McInnes.09.02.2018}.
We group the projected activations into $M$ clusters using k-means clustering and select the samples that are closest to the cluster centers. 
Additionally, we save the distance of the selected samples to the center of all projected samples of the current task, so that when a new task is added, we can discard samples of previous tasks that are furthest from the initial center. 

\section{Experiments}
We evaluate the sample selection methods for CSS in class- and in a domain-incremental setup.
For the experiments we follow the same training architecture and hyperparemters as \cite{Klingner2020, Kalb2021} and use ERFNet \cite{8063438} architecture, which has been pre-trained on ImageNet.
Adam optimizer is used in combination with a polynomial learning rate schedule that starts with a learning rate of $4\times 10^{-4}$ with power 0.9. 

\subsection{Class-Incremental Setup}
As most class-incremental benchmarks reuse the images in each task increment, they fail to account for the scenario where some of the classes will not reappear in the images of subsequent tasks.
As we want to focus in our experiments on this specific scenario, we adapt the Disjoint Cityscapes (DJ-CS) \footnote{The exact training splits can be found at \url{https://github.com/tobiaskalb/disjoint_cityscape_splits}.} split proposed in \cite{Kalb2021}. 
DJ-CS is divided into three distinct subsets, denoted as $S_1$, $S_2$ and $S_3$. 
In each of these subsets only a disjoint selection of classes are labeled. 
Furthermore, the classes \textit{truck}, \textit{bus}, \textit{motorcycle} only appear in the images of $S_2$, so that the information a teacher model can distill is limited.
Finally, only using naive replay is not sufficient for class-incremental CSS because in each image large parts remain unlabelled and the classes would never occur in the same image, Hence, the model would not be able to distinguish between classes of different subsets.
Therefore, we use replay in combination with CIL \cite{Klingner2020} in the class-incremental setting.

\subsection{Domain-Incremental Setup}
Domain-Incremental learning is a less challenging task than class-incremental learning \cite{Hsu18_EvalCL}, but as replay-based methods have shown to be very effective in this scenario \cite{Kalb2021}, we validate that the proposed approaches will also work in a setting in which only the underlying input distribution is changing. 
Therefore, we establish a domain incremental setting in which the segmentation model is first trained on BDD100k \cite{Yu.2020} and subsequently on Cityscapes \cite{Cordts.06.04.2016}. 
In the domain-incremental setting we use naive replay and the basic cross-entropy loss.

\begin{figure}[]
\centering
\includegraphics[width=\columnwidth]{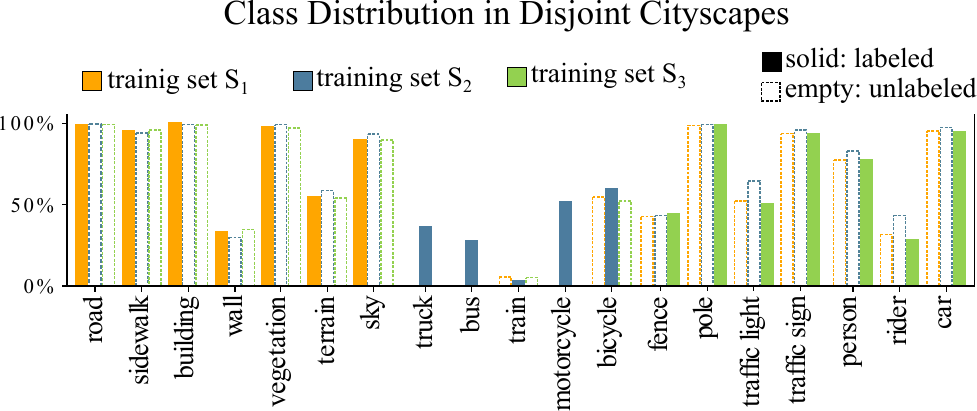} 
\caption{The training subsets of Disjoint Cityscapes and their respective class distributions. The solid bars indicate classes that are labeled in this subset, whereas the empty bars denote classes that are ignored. The classes \textit{truck}, \textit{bus} and \textit{motorcycle} only appear in the images of training set $S_2$.}
\label{fig:class-distribution-class-incremental}
\end{figure}

\section{Experimental Results}
\begin{figure}[]
\centering
\includegraphics[width=\columnwidth]{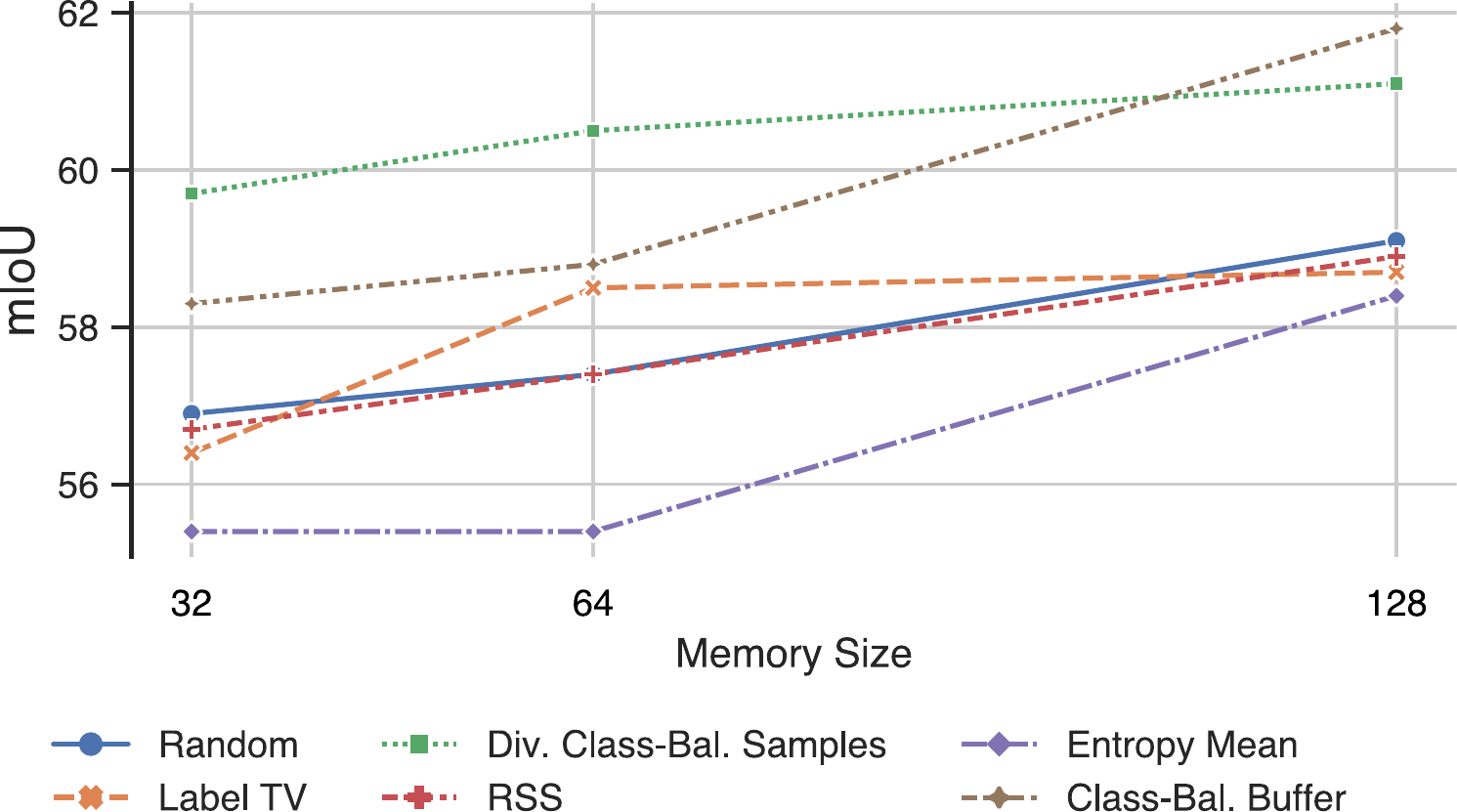}
\caption{Influence of the Memory Size M on the mIoU [\%] of different sample selection methods in the class-incremental setting.}
\label{fig:class_inc_memory}
\end{figure}

\begin{figure}[]
\centering
\includegraphics[width=\columnwidth]{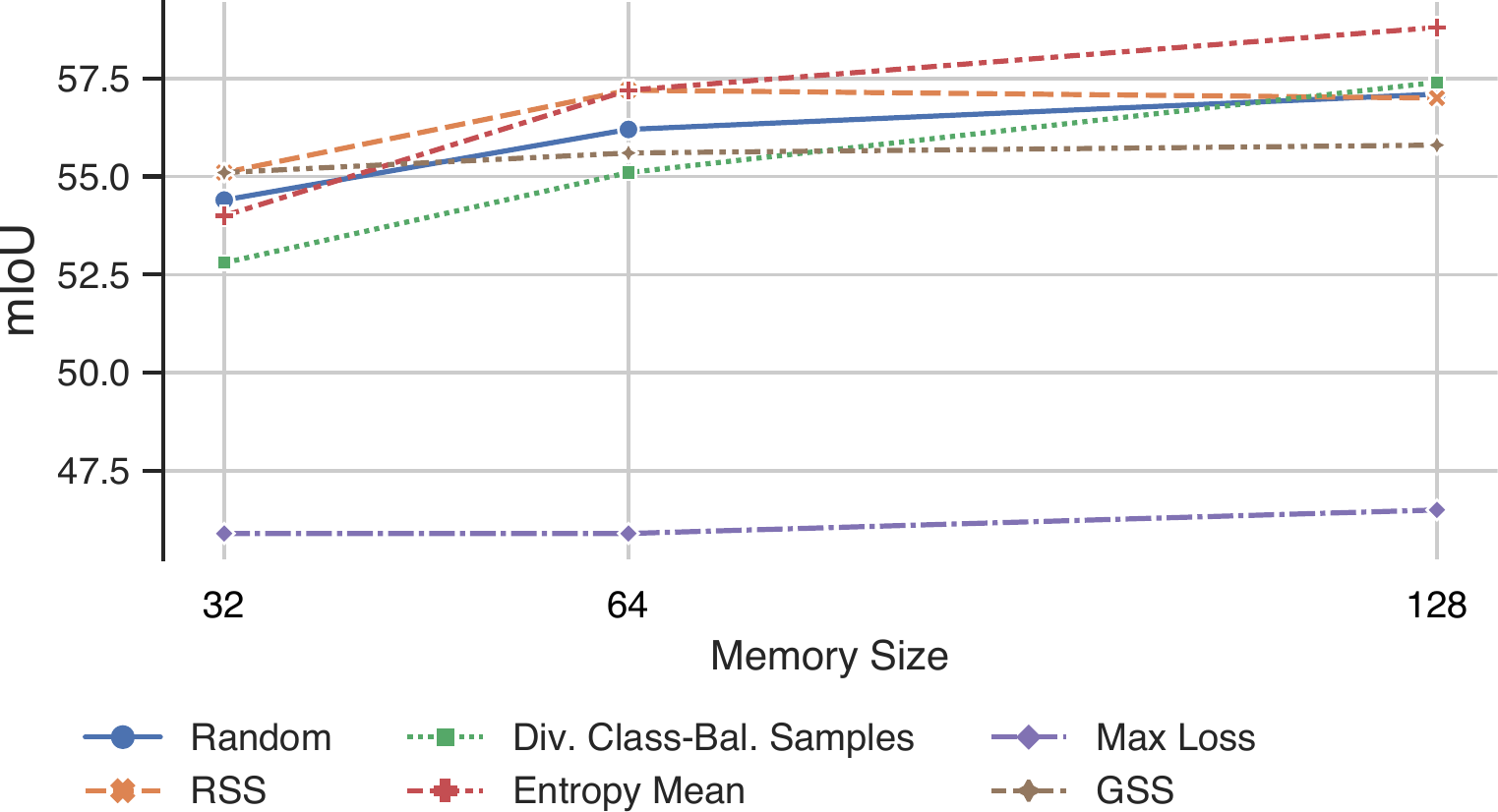}
\caption{Influence of the Memory Size M on the mIoU [\%] of different sample selection methods in the domain-incremental setting.}
\label{fig:domain_inc_memory}
\end{figure}

We evaluate the performance of each sampling method for replay-based learning on the class-incremental DJ-CS setup and the domain-incremental BDD100k to Cityscapes benchmark. 
We compare these methods to joint training on all tasks (Offline) and fine-tuning (FT), as respective upper and lower bound to the performance.
Additionally, in the class-incremental setting we use CIL \cite{Klingner2020} without replay as a baseline method.
In all our experiments we use a buffer size of 64. Results with buffer size 32 and 128 are shown in \cref{fig:class_inc_memory} and \cref{fig:domain_inc_memory}.
The results of the class-incremental and domain-incremental experiments are displayed in Tables \ref{tab:class_incremental_64}, \ref{tab:class_incremental_64_classes} and Table \ref{tab:domain_incremental_64}.

\subsection{Results on class-incremental learning}
We report the results of the Disjoint-Cityscapes experiments in \cref{tab:class_incremental_64}.
We observe that that FT performs well on the most recent task, but suffers from severe forgetting on previous tasks leading to an mIoU of 3.6\% when evaluated on all classes. 
CIL reduces forgetting considerably on previous tasks while adapting well to new classes. Although these results are outperforming FT by a large margin, they are still far from the joint model's accuracy.
Replay-based methods provide stable improvements on previous tasks compared to CIL, especially on $S_2$ that contains classes that do not appear in $S_1$ or $S_3$, in which the best performing method outperforms CIL by an absolute of 14.3\% mIoU. 
This demonstrates that Replay is most useful in a truly disjoint setting.
However, we also observe that using replay in the class-incremental setting leads to a drop in accuracy on new classes, meaning that replay might inhibit effectively learning new classes. 
It is possible that this effect arises from replay samples confusing the model, since new classes are labelled as background in replay samples.
Of all the sample selection methods only Min. Loss leads to significantly lower performance with an absolute difference of 5.22\% mIoU to the best performing method.
This can be attributed to the fact that Min. Loss tends to select easier samples containing only very few classes that take up large parts of the image, which leads to only 36.5\% mIoU on $S_2$. 
Overall, the different variants of entropy- and loss-based selection with the exception of Max. Loss underpeform compared to the remaining methods. 
Image-based methods that select samples according to their BRISQUE or Total Variation scores have a similar effect, since in the class-incremental setting, the holistic image quality or image complexity is less critical, as only small portions of the images are labelled.
Random Sample Selection achieves high accuracy on $S_1$ and $S_3$, but for $S_2$ it lacks an absolute of 9.0\% mIoU compared to the best performing selection method. This indicates that Random Selection often fails to select suitable samples for all the classes in $S_2$ and fails reduce the task-recency bias for classes of $S_2$.
On $S_2$, methods that aim to balance classes in the replay buffer achieve the highest mIoU, indicating that class-balancing is crucial for sample selection in the class incremental setting. By using LPIPS, we further improve the results for Class-Balanced Samples by discarding similar images from the buffer.
Diversity Balanced Samples achieved the highest overall mIoU, 58.7\% mIoU on $S_2$ and 60.5\% on Cityscapes validation set. For memory sizes 32 and 128, Class-Balanced Buffer performs better than Class-balanced samples, and both methods outperform the remaining methods for buffer sizes 32, 64, and 128
Results for different memory sizes are shown in \cref{fig:class_inc_memory}.
The class IoU of the most successful methods, RSS,  Min. Loss and Class-Balaneced Samples are shown in \cref{tab:class_incremental_64_classes}. 
Finally, \cref{fig:confmat_class_inc} shows that the Diversity Balanced Samples is effectively mitigating the bias of classes of $S_2$ to be miss-classified as classes of the most recent task.\\

\begin{table*}[]
    \centering
    \resizebox{0.85\textwidth}{!}
    {
    \begin{tabular}{l|cc|cccc}
     & \multicolumn{2}{l|}{Evaluation after Training on $S_1$, $S_2$} & \multicolumn{4}{l}{Evaluation after Training on $S_1$, $S_2$, $S_3$} \\ \hline
    Sample Method & mIoU\textsubscript{$S_1$} & mIoU\textsubscript{$S_2$} & mIoU\textsubscript{$S_1$} & mIoU\textsubscript{$S_2$} & \multicolumn{1}{l|}{mIoU\textsubscript{$S_3$}} & mIoU\textsubscript{Cityscapes} \\ \hline
    Offline  & - & - & 82.6 & 65.3 & \multicolumn{1}{l|}{79.2} & 69.0 \\
    Fine Tuning  & 20.7 & 61.8 & 6.1 & 12.2 & \multicolumn{1}{l|}{77.1} & 3.57 \\ 
    CIL \cite{Klingner2020} & 71.3 & 67.1 & 71.3 & 44.4 & \multicolumn{1}{l|}{82.4} & 55.8 \\ \hline
    Random Mean & 76.3 & 65.2 & 76.0 & 49.7 & \multicolumn{1}{l|}{81.7} & 57.4 \\
    Random Best of 10 & 75.4 & 63.4 & 75.2 & 56.6 & \multicolumn{1}{l|}{81.6} & 58.6 \\
    Random Worst of 10 & 76.1 & 63.0 & 75.1 & 42.4 & \multicolumn{1}{l|}{81.6} & 55.2 \\ \hline
    Min Loss & 73.4 & 62.2 & 73.9 & 43.3 & \multicolumn{1}{l|}{82.2} & 54.4 \\
    Max Loss & 76.5 & 67.8 & 76.3 & 48.3 & \multicolumn{1}{l|}{82.8} & 57.6 \\
    Mean Loss & 77.0 & 64.0 & 76.8 & 47.1 & \multicolumn{1}{l|}{82.3} & 58.4 \\
    Min Entropy & 75.7 & 72.6 & 75.7 & 48.2 & \multicolumn{1}{l|}{81.1} & 57.3 \\
    Max Entropy & 75.0 & 68.8 & 75.2 & 47.8 & \multicolumn{1}{l|}{81.6} & 56.2 \\
    Mean Entropy & 74.9 & 69.9 & 75.5 & 43.1 & \multicolumn{1}{l|}{80.9} & 57.2 \\
    Max BRISQUE & 76.2 & 68.3 & 56.6 & 49.3 & \multicolumn{1}{l|}{82.2} & 56.6 \\
    Total Variation (Label) & 75.6 & 71.1 & 75.7 & 36.9 & \multicolumn{1}{l|}{81.8} & 55.4 \\
    Total Variation (Image) & 74.6 & 61.6 & 74.5 & 49.5 & \multicolumn{1}{l|}{81.0} & 55.9 \\
    Ambivalent Classes & 75.1 & 63.3 & 75.0 & 49.6 & \multicolumn{1}{l|}{81.5} & 57.4 \\
    Class Bal. Buffer & 74.6 & 61.2 & 75.3 & 56.8 & \multicolumn{1}{l|}{81.6} & 58.8 \\
    Class Bal. Samples & 76.5 & 62.5 & 76.5 & 56.4 & \multicolumn{1}{l|}{82.6} & 59.8 \\
    Div. Bal. Samples $th=0.6$ & 77.5 & 66.6 & 77.0 & 58.7 & \multicolumn{1}{l|}{82.8} & 60.5 \\
    GSS \cite{{Aljundi.20.03.2019}} $cmp=5$ & 74.6 & 65.9 & 74.1 & 49.5 & \multicolumn{1}{l|}{80.0} & 56.6 \\
    RSS & 74.2 & 65.4 & 73.0 & 52.7 & \multicolumn{1}{l|}{81.7} & 57.4 \\
    \end{tabular}%
    }
    \caption{Results in mIoU [\%] of sample selection methods for the class-incremental setup with buffer size=64. Evaluation is run on the Cityscapes validation set using the indicated subsets shown in Fig. \ref{fig:class-distribution-class-incremental}, mIoU\textsubscript{$S_k$} denotes the mIoU only with respect to the current set of classes $S_k$.  }\label{tab:class_incremental_64}
\end{table*}
\begin{table*}
    \centering
	\resizebox{\textwidth}{!}{%
    \begin{tabular}{l|cccccccccccccccccccc}
     & \multicolumn{20}{l}{Evaluation after training on $S_1$, $S_2$, $S_3$} \\ \hline
    Sample method             & road 			& sidewalk 			& build 			& wall 				& veg. 				& terrain 			& sky 				& truck* 			& bus* 				& train 			& motor* 			& bicycle     		& fence 			& pole 				& TL 				& TS 				& person 			& rider 			& \multicolumn{1}{l|}{car} 					& mIoU 				\\ \hline
    Offline & 97.3 & 81.4 & 91.0 & 43.7 & 91.8 & 62.0 & 93.5 & 48.6 & 71.1 & 41.9 & 41.6 & 72.4 & 52.2 & 62.9 & 62.9 & 72.7 & 78.2 & 53.2 & \multicolumn{1}{l|}{93.0} & 69.0 \\
    CIL \cite{Klingner2020} & 95.6 & 69.8 & 86.9 & 19.8 & 88.7 & 47.4 & 86.6 & 7.1 & 17.5 & 26.8 & 0.0 & 58.7 & 45.4 & 60.7 & 62.3 & 70.8 & 76.4 & 52.2 & \multicolumn{1}{l|}{88.3} & 55.8 \\ 
    Random Mean & 96.0 & 73.5 & 87.1 & 29.7 & 89.5 & 52.3 & 89.0 & 14.3 & 24.2 & 24.6 & 13.4 & 58.3 & 45.3 & 57.8 & 57.9 & 68.3 & 73.9 & 47.8 & \multicolumn{1}{l|}{88.2} & 57.4 \\ \hline
    Div. Bal. Samples $th=0.6$ & 96.3 & 75.0 & 86.9 & 33.1 & 89.3 & 54.3 & 89.1 & 37.6 & 53.1 & 12.5 & 15.8 & 61.0 & 46.6 & 57.7 & 58.0 & 68.1 & 75.9 & 48.6 & \multicolumn{1}{l|}{90.8} & 60.5 \\ 
    Class Bal. Samples & 96.1 & 74.3 & 87.2 & 31.4 & 89.9 & 54.1 & 87.9 & 28.0 & 53.3 & 13.8 & 14.5 & 61.1 & 46.7 & 58.3 & 56.9 & 70.3 & 75.9 & 47.0 & \multicolumn{1}{l|}{90.4} & 59.8 \\ 
    RSS & 95.6 & 72.1 & 86.1 & 23.4 & 89.1 & 50.8 & 86.3 & 15.3 & 32.6 & 20.6 & 15.2 & 59.1 & 46.9 & 57.2 & 57.2 & 69.2 & 75.4 & 49.7 & \multicolumn{1}{l|}{88.8} & 57.4 \\ 
    Min Loss & 96.0 & 73.0 & 86.1 & 22.3 & 88.6 & 49.2 & 88.5 & 9.2 & 18.6 & 14.3 & 0.8 & 52.6 & 43.9 & 56.8 & 59.0 & 68.0 & 73.6 & 45.0 & \multicolumn{1}{l|}{87.5} & 54.4 \\ \hline
    \multicolumn{21}{r}{* appear only in $S_2$}
    \end{tabular}%
}
\caption{Results in IoU [\%] per class of sample selection methods for the class-incremental setup with buffer size 64. Evaluation is performed on all classes of the Cityscapes validation set after training on the entire task sequence}\label{tab:class_incremental_64_classes}
\end{table*}

\subsection{Results on domain-incremental learning}
Next up, the methods are also evaluated in domain-incremental settings. The results are summarized in \cref{tab:domain_incremental_64}.
In the domain-incremental setting, forgetting is less severe than in the class-incremental setting. 
Furthermore, the sample selection method does not have as much impact on accuracy as the class-incremental setting, except for Max. Loss, which produces the lowest mIoU.
The reason for the lackluster performance is that Max. Loss selected almost exclusively samples of BDD100k that contain erroneous labels, that result in a high loss. 
The selection by highest loss also shows that bad sample selection can inhibit learning of the new domain severely as it drops an absolute of 2.48\% mIoU on Cityscapes compared to random sample selection.
Furthermore, class-balancing methods do not achieve better accuracy than random selection, suggesting that in the domain-incremental setting class-balancing is of less importance.
On the other hand, model-based methods such as Mean Loss, Mean Entropy and Representation-based Sample Selection (RSS) are more effective in the domain-incremental scenario as in the class-incremental scenario. 
\cref{fig:cka_sim} shows the CKA similarity score \cite{Kornblith2019} of the intermediate representations of the model before and after training on the second task. We observe that the higher the similarity score, the better the method is suited to prevent forgetting. 
RSS achieves the highest similarity score in every layer. We attribute this to the fact that samples are chosen to approximate the distribution of the internal representations of the model, which stabilizes the representations of the network.
This suggests that the internal representations of the model play an important role for sample selection, particularly in the domain incremental setting.
\cref{fig:domain_inc_memory} shows that with decreasing memory size the sample selection method has a much bigger impact on the final performance. Meaning that when memory is scarce, sample selection becomes even more critical.
\addtolength{\textheight}{-0.5cm} 
\begin{table}[]
    \centering
    \resizebox{0.97\linewidth}{!}{
    \begin{tabular}{l|ccc}
     & \multicolumn{3}{c}{Training on BDD and Cityscapes} \\ \hline
    Sample Method & mIoU\textsubscript{BDD} & mIoU\textsubscript{Cityscapes} & mIoU\textsubscript{Average} \\ \hline
    Offline & 57.2 & 65.0 & 61.1 \\
    FT  & 29.6 & 67.0 & 48.3 \\ 
    Random Mean & 44.0 & 68.4 & 56.2 \\
    Random Best of 10 & 44.4 & 69.3 & 56.9 \\
    Random Worst of 10 & 42.3 & 68.0 & 55.2 \\\hline
    Min Loss & 39.9 & 69.2 & 54.6 \\
    Max Loss & 25.4 & 66.4 & 45.9 \\
    Mean Loss & 46.2 & 67.0 & 56.6 \\
    Min Entropy & 41.4 & 68.4 & 54.9 \\
    Max Entropy & 36.3 & 67.0 & 51.7 \\
    Mean Entropy & 45.0 & 69.4 & 57.2 \\
    BRISQUE & 42.1 & 66.5 & 54.3 \\
    Total Variation (Label) & 39.5 & 66.1 & 52.8 \\
    Total Variation (Image) & 44.4 & 67.3 & 55.9 \\
    Ambivalent Classes & 42.2 & 67.6 & 54.9 \\
    Class Bal. Buffer & 39.8 & 68.2 & 54.0 \\
    Class Bal. Samples & 40.6 & 68.6 & 54.6 \\
    Div. Bal.Samples $th=0.6$ & 41.7 & 68.6 & 55.1 \\
    GSS \cite{{Aljundi.20.03.2019}} $cmp=5$ & 44.9 & 68.6 & 55.6 \\
    RSS & 45.5 & 69.0 & 57.2 \\
    \end{tabular}
    }
    \caption{Results in mIoU [\%] of sample selection methods for domain-incremental setting with buffer size = 64. Evaluation is run after training on both datasets.}\label{tab:domain_incremental_64}
\end{table}

\section{Conclusion}
We evaluated a diverse set of sample selection methods for replay-based class- and domain-incremental semantic segmentation. 
We demonstrate the necessity of replay for scenarios in which classes do not reoccur in subsequent tasks, as in this scenario approaches based solely on knowledge distillation fail to retain knowledge for all classes.
Furthermore, we illustrate the importance of sample selection in CSS. In the class-incremental setting, class-balancing sample selection methods have proven to be the most effective, while methods based on loss statistics, entropy or the internal representations of the model fail for non-reoccurring classes. In the domain incremental setting, however, we observe that the most effective sample selection methods help to reduce the internal representational shift, e.g., by approximating the distribution of their internal representation using RSS or by selecting samples that have a median entropy.
For edge applications, where only limited memory is available, we unveil that necessity of effective sample selection in order to guarantee the diversity of data utilized for continual learning.

\section*{ACKNOWLEDGMENT}
The research leading to these results is funded by the German Federal Ministry for Economic Affairs and Climate Action within the project “KI Delta Learning“ (Förderkennzeichen 19A19013T). The authors would like to thank the consortium for the successful cooperation.

\bibliographystyle{IEEEtran}
\bibliography{IEEEabrv,egbib}

\end{document}